%% file: main.tex
\documentclass[times,twocolumn,final]{elsarticle}

\usepackage{framed,multirow}
\usepackage{url}
\usepackage{xcolor}
\definecolor{newcolor}{rgb}{.8,.349,.1}
\usepackage{amsmath,amssymb,amsfonts}
\usepackage{algorithm,algorithmic}
\usepackage{graphicx}
\usepackage{textcomp}

\usepackage{grffile}
\usepackage{longtable}
\usepackage{wrapfig}
\usepackage{rotating}
\usepackage[normalem]{ulem}
\usepackage{capt-of}
\usepackage{hyperref}
\usepackage{float}
\usepackage{textcomp}

\usepackage{lipsum}
\usepackage{pifont}
\usepackage{bm}
\usepackage{booktabs}
\usepackage{mwe}
\usepackage{arydshln}
\input{macros.tex}

\begin{document}

\begin{frontmatter}
\title{A Positive/Unlabeled Approach for the Segmentation of Medical Sequences using Point-Wise Supervision}
\author{Laurent Lejeune}
\ead{me@lejeunel.org}
\author{Raphael Sznitman}

\address{Artificial Intelligence in Medical Imaging, ARTORG Center, University of Bern, Murtenstrasse 50, 3008 Bern, Switzerland}
\input{abstract.tex}
\end{frontmatter}

\input{intro.tex}
\input{rel_works.tex}
\input{methods.tex}
\input{exp.tex}
\input{conclusion.tex}

\section*{Acknowledgments}
This work was supported in part by the Swiss National Science Foundation Grant 200021 162347 and the University of Bern.

\bibliographystyle{model2-names.bst}\biboptions{authoryear}
\bibliography{refs}

%\printbibliography
\end{document}

%% file: macros.tex
\newcommand{\SSnnPU}   {{\bf SSnnPU}}
\newcommand{\AlterEstPU}   {{\bf AlterEstPU}}
\newcommand{\KSPTrack}{{\bf KSPTrack}}
\newcommand{\SSnnPUKSP}{{\bf SSnnPU}+{\bf KSPTrack}}
\newcommand{\SSnnPUTrue}{{\bf nnPUTrue}+{\bf KSPTrack}}
\newcommand{\SSnnPUConst}{{\bf nnPUConst}+{\bf KSPTrack}}
\newcommand{\MaxSP}{{\bf Max. SP}}

\newcommand{\eg}{\emph{e.g., }}
\newcommand{\ie}{\emph{i.e., }}

\newcommand{\red}[1]{#1}

%% file: abstract.tex
\begin{abstract}

The ability to quickly annotate medical imaging data plays a critical role in training deep learning frameworks for segmentation. Doing so for image volumes or video sequences is even more pressing as annotating these is particularly burdensome. To alleviate this problem, this work proposes a new method to efficiently segment medical imaging volumes or videos using point-wise annotations only. This allows annotations to be collected extremely quickly and remains applicable to numerous segmentation tasks.
Our approach trains a deep learning model using an appropriate Positive/Unlabeled objective function using sparse point-wise annotations. While most methods of this kind assume that the proportion of positive samples in the data is known a-priori, we introduce a novel self-supervised method to estimate this prior efficiently by combining a Bayesian estimation framework and new stopping criteria. Our method iteratively estimates appropriate class priors and yields high segmentation quality for a variety of object types and imaging modalities. In addition, by leveraging a spatio-temporal tracking framework, we regularize our predictions by leveraging the complete data volume. We show experimentally that our approach outperforms state-of-the-art methods tailored to the same problem.

\end{abstract}

\begin{keyword}
Transductive Learning, Positive-Unlabeled learning, Semantic segmentation, Point-wise supervision
\end{keyword}

%%% Local Variables:
%%% mode: latex
%%% TeX-master: "main"
%%% End:

%% file: intro.tex
\section{Introduction}
\label{sec:intro}
Modern machine learning methods for semantic segmentation have shown excellent performances in recent years across numerous medical domains. These advances have been spearheaded by new deep learning based approaches that leverage large amounts of images and annotations. Unfortunately however, generating substantial quantities of segmentation annotations for medical imaging remains extremely burdensome. This is in part due to the need for domain specific knowledge in clinical applications, but also due to the fact that many medical imaging modalities are often 3-dimensional (\eg CT, MRI) or video based (\eg endoscopy, microscopy, etc. ). The latter point greatly increases the necessary time and effort to manually segment even just a few volumes or videos.
\begin{figure*}[t]
\centering
\includegraphics[width=0.99\textwidth]{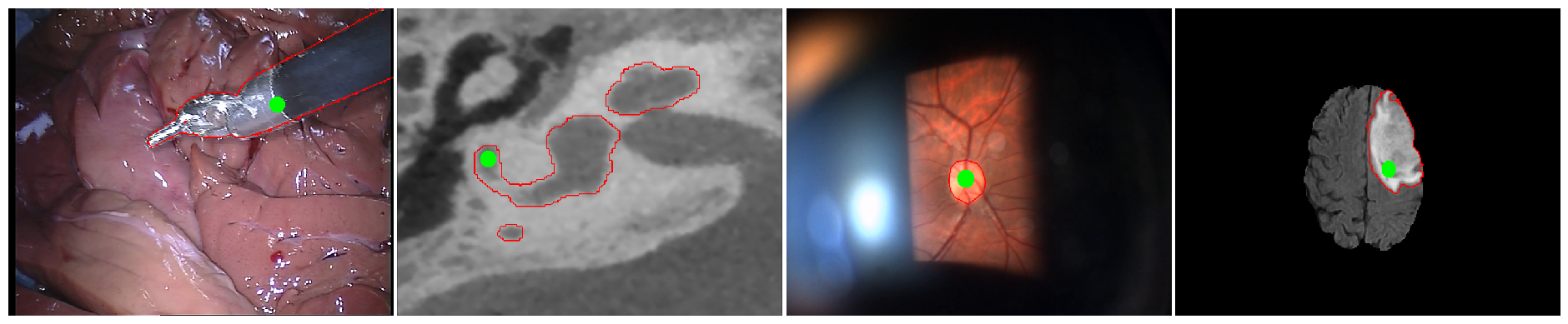}
\caption{Examples of image frames in different imaging modalities with different objects of interest to segment. In each example, a 2D point annotation is show in green and the complete pixel-wise groundtruth segmentation is shown in red. Applications shown are (from left to right): Video frame of a surgical instrument during minimally invasive surgery, a single slice from a CT scan depicting a human cochlea, video frame from a slitlamp examination of the optic nerve, brain slice from an MRI scan showing a tumor.}
\label{fig:intro}
\end{figure*}

To overcome this important bottleneck, semi-supervised learning in medical imaging has been an active research area. Sub-domains have included, active learning~\citep{sener18, KonSznFua15}, self-supervised methods~\citep{chen19, jamaludin17}, or crowd-sourcing based annotating~\citep{salvador13, heim18}, all of which have variants tailored for segmentation tasks. Broadly speaking, the core idea in each of these is to maximize the value of individual manually generated annotations for subsequent training of segmentation models. One sub-type of semi-supervised methods, known as {\it transductive learning}, considers the case where all data points are initially given and the labels of certain data points must be inferred from a subset of known annotated data points. As such, it is closely related to belief propagation~\citep{knobelreiter2020belief} and unsupervised learning. 

We thus consider inferring a complete segmentations of a given 3D volume or video sequence from limited user-provided annotations, as a transductive learning problem. That is, we wish to segment all pixels of a given data volume, whilst only having access to a few partial locations being annotated.  \red{Unlike traditional graphcut based methods~\citep{boykov01} that do so by using both positive and negative samples (\ie PN learning), our focus is on cases where only positive samples are accessible.} With some application-specific solutions previously developed~\citep{vilarino2007,khosravan16}, we follow the line of~\citep{bearman16,lejeune17,lejeune18}, aiming for agnostic solutions capable of working for different unknown object of interest (shape, appearance, motion, etc. ), as well as different imaging modality (MRI, CT-scan, video, etc.). Fig.~\ref{fig:intro} illustrates a number of different such scenarios and highlights the broad range appearances and settings considered here.

To further reduce the annotation burden, we choose to make use of sparse point-wise annotations to indicate pixels that belong to the object of interest in given data volumes. As observed in~\cite{bearman16}, point-wise supervision are extremely easy to collect and very reliable. While a manual segmentation may take on average 239 seconds for a single PASCAL VOC image, the corresponding multi-class point-wise annotations only requires 22 seconds. In this work, we further restrict the point-wise annotations to only be available on the object of interest, and not on the background regions of the image. Additionally, point-wise annotations can be provided by manual clicks~\citep{ferreira12,bearman16}, or using a gaze tracker~\citep{vilarino2007,khosravan16,lejeune17,lejeune18}. While this convenience and speed gain comes at the cost of extremely sparse annotations in contrast to full semantic segmentations, methods to generate full segmentations in this setting have shown some promising performances in medical imaging. The present work follows this specific line of research.

Our goal is thus to generate a sequence of binary segmentation masks for an object of interest present in a volume or video using point-wise annotations that only specify the object. That is, the annotations only provide explicit information as to the location of the object of interest and no information regarding the background is known. To do this, we would ideally like to train a function to learn from annotated locations, take into account unlabeled locations and infer their labels. While it would appear that using neural networks to do so would be the obvious choice, doing so is challenging because positive samples are given by annotations (\ie locations on the object of interest throughout the data), yet no explicit negative samples are available. Instead, a large unlabeled set of samples is accessible without knowing which of these is positive or negative. This problem setting is known as P(ositive)-U(nlabeled) learning~\citep{li03,li05,duplessis14,duplessis15} and lies at the heart of this work.

We thus introduce a novel PU learning method that allows for high segmentation performances from point-wise annotations in a transductive learning setting. Our approach leverages a non-negative unbiased risk estimator to infer the likelihood of the object presence throughout the data. Specifically, we use this estimator as an objective function to train a deep learning model in a transductive setup. As the estimator requires accurate class-priors to be effective, we introduce a self-supervised strategy to estimate the proportion of positive samples on each frame using a recursive bayesian approach within our training procedure. Our method has the benefit of only needing a single upper-bound initialization value while allowing per-frame estimates to be computed. We further combine the latter estimates with a multi-path tracking framework that explicitly leverages the spatio-temporal relations of over-segmented regions. This allows the output of our model to be regularized throughout the data volume. We show experimentally that our pipeline brings important performance gains over state-of-the-art methods across a broad range of image modalities and object types.

%%% Local Variables:
%%% mode: latex
%%% TeX-master: "main"
%%% End:

%% file: rel_works.tex
\section{Related Works}
\label{sec:rel_works}

%The present paper combines theoretical contributions related to semi-supervised learning, of which Positive/Unlabeled learning is a sub-category, as well as self-%supervised learning.
%From an application point of view, we contribute to the field of sparse point-wise supervision for segmentation.
%We now give an overview of the state-of-the-art related to the present contributions.

In this section we provide an overview of some of the most relevant related works to the method presented here.

\textbf{Positive/Unlabeled learning} considers the learning setting where only a subset of the positive samples are labeled, while the unlabeled set contains both positive and negative samples.
Early methods focused on iteratively sampling confident negatives from the unlabeled set using a classifier, while re-training the same classifier using these~\citep{li03,liu03,li05}.
In~\cite{lee03}, the authors propose a reweighing scheme applied to the unlabeled samples, which allows the use of traditional supervised classifiers.
As the latter approach heavily relies on appropriate weights,~\cite{elkan08} instead chose to duplicate unlabeled samples into positive and negative samples with complementary weights, an approach called unbiased PU learning. More recently, a general-purpose unbiased risk estimator for PU learning was presented by~\cite{duplessis15} which allows for convex optimization in the PU setting. As a follow-up to the latter,~\cite{kiryo17} noted that modern expressive models, such as Deep Neural Networks, induce negative empirical risks through overfitting of the positives, which they propose to fix by introducing a non-negative unbiased risk estimator. We detail this method in the next section as we build directly from this method. Beyond the works mentioned here, we invite the reader to a more complete review of the topic in~\cite{bekker20}, \red{as many new methods have now focused on specific settings such as biased-negative samples~\citep{hsieh19} or temporal shifts in the positive class~\citep{akujuobi2020}.}

\textbf{Class-prior estimation} is tightly related to the state-of-the-art PU learning approaches that design risk estimators relying on knowing or estimating the density of positive and negative samples. In~\cite{duplessis14}, the authors suggest to partially match the class-conditional densities of the positive class to the input samples using the Pearson divergence criteria. In~\cite{christoffel16}, the same approach is improved by considering a general divergence criteria along with a $L_{1}$ distance regularization, which diminishes the problem of over-estimation. In~\cite{bekker18}, a tree induction scheme is introduced to estimate the probability of positive samples as a proxy task~\citep{scott09}. Similar to the self-supervised estimation of class-priors proposed in the present work,~\cite{kato18} combines the non-negative unbiased risk of~\cite{kiryo17} with an iterative update of class-priors. In particular, they devise an update rule inspired by the Expectation-Maximization algorithm and iterate until convergence.

\textbf{Point-wise supervision} was first applied in the context of medical image analysis in~\cite{vilarino2007}, where a Support Vector Machine was used to classify patches. Using a graph approach,~\cite{khosravan16} constructed saliency maps as the input of a Random-Walker to segment CT volumes. More generally,~\cite{bearman16} train a CNN using a loss function that includes an object-prior defined by the user-provided points. The most relevant methods to the present work are those of~\cite{lejeune17} and~\cite{lejeune18}. In the former, a classifier is learned to segment various kinds of medical image volumes and videos in a PU setting using a loss function that leverages the uncertainties of unlabeled samples. As a follow-up,~\cite{lejeune18}, formulated the same problem as a multi-path optimization problem. In particular, a simple object model formulated in a PU setting provides costs of selecting superpixels. The multi-path framework showed good performances in inferring structures that were non-concave (\eg a doughnut shape).

%%% Local Variables:
%%% mode: latex
%%% TeX-master: "main"
%%% End:

%% file: methods.tex
\section{Methods}
\label{sec:methods}

The goal of our method is to generate a segmentation mask for an object of interest in each frame of a single image volume or video sequence using only point-wise annotations and without knowing the image type or object of interest beforehand.

To do so, we propose a novel approach within the Positive-Unlabeled learning setting, that learns the segmentation of the object identified by the point-wise annotations. Our method makes use of the non-negative risk estimator introduced in~\cite{kiryo17}, which heavily relies on knowing the proportion of positive samples in the data. While unavailable in our setting, we introduce a novel self-supervised method to estimate this key value via an iterative learning procedure within a Bayesian estimation framework. We then devise a stopping condition to halt training at an appropriate point. Last, a spatio-temporal tracking framework is applied to regularize the output of our approach over the complete volume information.

We describe our approach now in more detail. In the following subsection, we introduce the Positive-Unlabeled learning framework and its non-negative risk estimator. We then describe in 
Sec.~\ref{sec:pi_estim} our self-supervised approach to learn effective class priors. Last, we detail how we leverage the spatio-temporal regularizer, and provide our implementation details
in Sec.~\ref{sec:tracking} and Sec.~\ref{sec:implementation}, respectively.
\input{method_nnpu.tex}
\input{method_bayesian.tex}
\input{method_tracking.tex}
\input{method_train_details.tex}

%%% Local Variables:
%%% mode: latex
%%% TeX-master: "main"
%%% End:

%% file: method_nnpu.tex
\subsection{Non-negative Positive-Unlabeled learning}
\label{sec:nnpu}

We first briefly introduce and formulate Non-negative Positive-Unlabeled learning~\cite{kiryo17} in the context of semantic segmentation.

Traditional supervised learning, which we denote Positive/Negative learning (PN), looks to build a model \(f_{\theta}: I \mapsto [0;1]^{W \times H}\), where \(\theta\) is a set of model parameters, \(I\) is the input image with width $W$ and height $H$. Letting \(\bm{\mathcal{I}} = \{\mathcal{I}^i\}_{i=1}^{N}\) be the set of $N$ input images corresponding to an image volume or video sequence, 
each image $\mathcal{I}^i$ is composed of pixels, $\mathcal{X}^i=\mathcal{X}_p^i \cup \mathcal{X}_n^i$, where $\mathcal{X}_p^i$ and $\mathcal{X}_n^i$ denote the positive and negative pixels, respectively. We denote $\pi^i \in (0,1)$ as the proportion of positive pixels in image $i$, and $\bm{\pi} = \{\pi^i\}_{i=1}^{N}$ as the set of such proportions over all frames. 
%To simplify the notation, we write $f_{\theta}(x) = f_{\theta}(\mathcal{I})\big|_x$ as theresponse of $f_{\theta}$ on image $\mathcal{I}$ averaged on the region defined by superpixel $x$.

Training $f$ to optimize $\theta$ can then be computed by minimizing the empirical risk of the form,
\begin{equation}
R_{pn}=\sum_{i=1}^{N} \left[ \frac{\pi^i}{|\mathcal{X}_p^i|}\sum_{x \in \mathcal{X}_p^i}\ell^+(f_{\theta}(x)) \right.+
\left. \frac{1-\pi^i}{|\mathcal{X}_n|} \sum_{x \in \mathcal{X}_n^i}\ell^-(f_{\theta}(x)) \right].
  \label{eq:pn}
\end{equation}
\noindent 
A popular choice for $\ell$ is the logistic loss, for which \(\ell^+(z)=\log(1+ e^{-z})\), \(\ell^-(z)=\log(1+e^{z})\) are the positive and negative entropy loss terms, respectively. In which case, Eq. \eqref{eq:pn} is the Balanced Cross-Entropy loss (BBCE).

Conversely, computing Eq.~\eqref{eq:pn} is infeasable in a PU settting, as neither $\bm{\pi}$ nor $\bm{\mathcal{X}}_{n}$ are known in advance. Instead, we have a set of unlabeled samples $\bm{\mathcal{X}}_{u}$ that contain both positives and negatives. As suggested in~\cite{duplessis15}, the negative risk (\ie the second term of Eq.~\eqref{eq:pn}) can however be re-written in terms of $\bm{\mathcal{X}}_{p}$ and $\bm{\mathcal{X}}_{u}$ as,

\begin{multline}
  \label{eq:nnpu}
R_{pu}=\sum_{i=1}^{N}\Biggl[ \frac{\pi^{i}}{|\mathcal{X}^{i}_{p}|}\sum_{x \in \mathcal{X}^{i}_p}\ell^+(f_{\theta}(x)) + \\
\Biggl( \frac{1}{|\mathcal{X}^{i}_{u}|}\sum_{x \in \mathcal{X}^{i}_u}\ell^-(f_{\theta}(x)) -
\frac{\pi^{i}}{|\mathcal{X}^{i}_{p}|}\sum_{x \in \mathcal{X}^{i}_p}\ell^-(f_{\theta}(x)) \Biggr) \Biggr].
\end{multline}
This is achieved by observing that $p(x) = \pi p(x|Y=1) + (1-\pi) p(x|Y=-1)$ and that the negative risk can be expressed as $ (1-\pi) \mathbb{E}_{X \sim p(x|Y=-1)}\left[\ell^-(f_\theta(X)) \right] =
\mathbb{E}_{X \sim p(x)}\left[\ell^-(f_\theta(X)) \right] - \pi \mathbb{E}_{X \sim p(x|Y=+1)}\left[\ell^-(f_\theta(X)) \right]$. In the case of expressive models such as Neural Networks, minimizing the objective of Eq.~\eqref{eq:nnpu} using stochastic gradient descent on mini-batches of samples tends to overfit to the training data, by driving the negative risk, (\ie the bottom term of Eq.~\eqref{eq:nnpu}) to be negative.

To circumvent this,~\cite{kiryo17} proposed to perform gradient ascent when the negative risk of a mini-batch is negative using the following negative risk,
\begin{multline}
  \label{eq:neg_risk}
R_{i}^{-}=\sum_{i=1}^N\Biggl(
 \frac{1}{|\mathcal{X}^{i}_{u}|}\sum_{x \in \mathcal{X}^{i}_u}\ell^-(f_{\theta}(x)) - 
\frac{\pi^{i}}{|\mathcal{X}^{i}_{p}|}\sum_{x \in \mathcal{X}^{i}_p}\ell^-(f_{\theta}(x)) \Biggr).
\end{multline}

Thus the complete training procedure for the PU setting with deep neural networks is described in Alg.~\ref{alg:sgdnnpu}. Specifically, when \(R_{i}^{-} < 0\), gradient ascent is performed by setting the gradient to \(-\nabla_\theta R_{i}^{-}\). 
%\algnewcommand\algorithmicforeach{\textbf{for each}}
%\algdef{S}[FOR]{ForEach}[1]{\algorithmicforeach\ #1\ \algorithmicdo}
\begin{algorithm}[H]
\caption{Non-negative PU learning}
\label{alg:sgdnnpu}
\begin{algorithmic}[1]
\REQUIRE{$f_\theta$: Prediction model \newline
  $\bm{\mathcal{I}}$: Set of images \newline
  $\bm{\mathcal{X}_p}, \bm{\mathcal{X}_u}$: Positive and unlabeled samples \newline
  $\bm{\pi}$: Set of class priors \newline
  $T$: Number of epochs}
\FOR{\texttt{epoch} $\gets 1$ to $T$}
\STATE Shuffle dataset into $N_{b}$ batches
    \FOR{$i \gets 1$ to $N_{b}$}
     \STATE Sample next batch to get $\mathcal{I}^i$, $\mathcal{X}_p^i$, $\mathcal{X}_u^i$, $\bm{\pi}^i$
	\STATE Forward pass $\mathcal{I}^i$ in $f_\theta$
	\STATE Compute risks as in Eq. \ref{eq:nnpu} and \ref{eq:neg_risk}
      \IF{$R_{i}^{-} < 0$}
          \STATE Do gradient ascent along $\nabla_\theta R_{i}^{-}$
      \ELSE
          \STATE Do gradient descent along $\nabla_\theta R_{pu}$
      \ENDIF
  \ENDFOR
\ENDFOR
\end{algorithmic}
\end{algorithm}

Critically, $\bm{\pi}$ plays an important role in Eq.~\eqref{eq:nnpu} and Eq.~\eqref{eq:neg_risk}. While~\cite{kiryo17} assumes that the class prior is known and constant across all the data, this is not the case in many applications, including the one at the heart of this work. In particular, the class prior here is specific to each frame of the image data available (\ie $\pi^i$) as each frame may have different numbers of positives (\eg the object may appear bigger or smaller). In addition, we show in our experimental section that setting the class prior in naive ways leads to low performance levels. In the subsequent subsection, we hence introduce a novel approach to overcome this limitation.

%%% Local Variables:
%%% mode: latex
%%% TeX-master: "main"
%%% End:

%% file: method_bayesian.tex
\subsection{Self-supervised Class-priors Estimation}
\label{sec:pi_estim}
Instead of fixing the values of $\bm{\pi}$ before training as in~\cite{kiryo17}, we instead propose to refine all values iteratively during training.  
Our approach, which we refer to as Self-Supervised Non-Negative PU Learning (\SSnnPU), will start with a large-upper bound for $\pi_i$ and will progressively reduce the estimates at each epoch of our training scheme until a stopping criterion is reached. That is, we will optimize the function $f_\theta$ one epoch at a time using Alg.~\ref{alg:sgdnnpu}, and then use the intermediary model to help estimate the class priors.

However, deriving class prior estimates from partial models (\ie trained with few epochs) yields very noisy estimates with large variances. Hence, we propose to use a Bayesian framework to estimate the class priors in a recursive fashion by establishing a state space and observation model and inferring the class priors. \red{This is motivated by the fact that most PU learning methods developed so far~\citep{bekker20} rely on the Selected Completely At Random (SCAR) assumption to model positive samples. In our case, this is not the case however as positive samples are highly correlated given taht they correspond to pixels in images and thus have strong correlations with other positive samples}.  We now describe our approach in more detail by first formalizing the state and observation models, and we describe our recursive Bayesian estimate and stopping conditions thereafter. Our final training algorithm is summarized in Alg.~\ref{alg:prior_estim}.

\subsubsection{State and Observation models}
Recall $\bm{\pi}_k=\{\pi_{k}^{i}\}_{i=1}^N$ to be the true class prior of a sequence of $N$ frames. While we wish to know $\bm{\pi}$, our method 
will compute values $\bm{\hat\pi}=\{\hat\pi^{i}\}_{i=1}^N$ as the best approximation to $\bm\pi$. At the same time, our prediction model after $k$ training epochs, denoted $f_{\theta_k}$, can also provide partial information to the value of $\bm\pi$. Specifically, by evaluating $f_{\theta_k}$ on all samples $x \in \mathcal{X}^i$, we can estimate a noisy observation of $\pi^i$ by computing the expected value over $f_{\theta_k}$,
\begin{equation}
  \label{eq:observ}
\rho_{k}^{i} = \mathbb{E}_{x \in \mathcal{X}^{i}}[f_{\theta_{k}}(x)^{\gamma}].
\end{equation}
\noindent 
Here, $\gamma > 1$ is a correction factor that mitigates variations in the expectation at different epoch values. That is, we wish that our prediction model slightly over-segment the object of interest so to over-estimate the frequencies of positives. This is because we wish to progressively decrease $\bm{\hat\pi}$ from its initial value $\hat \pi_{0}$ by using $\bm{\rho}_{k}$ as observations. 

To do this, we denote $\bm{\pi_k} $ and $\bm{\hat\pi_k}$ to be true and inferred class priors after epoch $k$. While the value $\bm{\pi_k} $ is the same for all values of $k$, we include this notation at this stage to define the following linear state observation model we will use to infer $\bm{\hat\pi_k}$,
\begin{align}
\bm{\bm{\pi}}_{k+1} &= g(\bm{\pi}_{k}, L) - u_{k}\mathbf{1}_{N} + \mathcal{N}(0,Q),\label{eq:trans_fn}
\end{align}
\noindent
where $\bm{\bm{\rho}}_{k} \sim \mathcal{N}(\bm{\pi}_{k},R)$,
$\bm{\pi}_{0}  \sim \mathcal{N}(\hat\pi_{0},S)$, and $Q$, $R$, and $S$ are the transition, observation, and initial covariance matrices, respectively. The function $g(\cdot, L)$ is a moving average filter of length $L$ with a Hanning window to impose a frame-wise smoothness. For convenience, we write $\mathbf{1}_{N}$ for a vector of length $N$ taking values of $1$, and the term $u_{k}$ is the control input,
\begin{equation}
u_{k} = u_{0} + (u_{T} - u_{0})\frac{k}{T},
\end{equation}
\noindent
where $u_{0}$ and $u_{T}$ are two scalars such that $u_{T} > u_{0}$.
The control input therefore induces a downward acceleration on the states and imposes a ``sweeping'' effect on the latter, which allows in principle the range $[0; \hat \pi_{0}]$ to be explored. Last, we also set $\forall i, \hat \pi^i_0 = \pi_{max} >> \pi^{i} $.

%\begin{align}
%\bm{\bm{\tilde\pi}}_{k+1} &= g(\bm{\tilde\pi}_{k}, L) - u_{k}\mathbf{1}_{N} + \mathcal{N}(0,Q), \quad \tilde\pi_{k}^{i} \in [0; \pi_{{max}}] \label{eq:trans_fn}\\
%\bm{\bm{\rho}}_{k} &= \bm{\tilde\pi}_{k} + \mathcal{N}(0,R) \label{eq:proc_fn} \\
%\bm{\tilde\pi}_{0}&=\hat \pi_{0} + \mathcal{N}(0,S) \label{eq:init_fn}
%\end{align}

%\subsubsection{Observation model}
%\label{sec:obs_model}
%We are now interested in inferring robust obervations from our prediction model.
%For clarity, we write $f_{\theta_{k}}$ for our prediction model trained for $k$ epochs.
%Let $f_{\theta_{k}}(x)$ the prediction given by $f_{\theta_{k}}$ on superpixel $x$ at iteration $k$.
%We therefore compute $\rho_{k}^{i}$, the observation at time $k$ on frame $i$ as:
%
%\begin{equation}
%  \label{eq:observ}
%\rho_{k}^{i} = \mathbb{E}_{x \in \mathcal{X}^{i}}[f_{\theta_{k}}(x)^{\gamma}]
%\end{equation}
%
%With $\gamma > 1$ a correction factor.
%This correction is justified as follows: As early iterations use class-priors that are above the true values,
%our prediction model tends to over-segment the object of interest and therefore over-estimate the frequencies of positives.

\begin{algorithm}[t]
\caption{Self-supervised Non-Negative PU Learning}
\label{alg:prior_estim}
\begin{algorithmic}[1]
\REQUIRE{$\hat \pi_{0}$: Upper-bound on class-priors \newline
  $T$: Number of epochs \newline
  $f_\theta$: Foreground prediction model}
\ENSURE {Optimal estimate of class-prior $\bm{\hat{\pi}}^{*}$\newline}
\STATE $k = 0$
\STATE $ {\hat{\pi}^i}_{0} = \pi_{max}, \forall i$
\WHILE {Stopping condition not satisfied}
    \STATE Optimize $f_\theta$ for $1$ epoch using Alg.~\ref{alg:sgdnnpu} and $\bm{\hat{\pi}}_{k}$
	%\STATE Forward pass all images to get $\bm{y}_k$
    \STATE Compute observations $\bm{\rho}_k$ as in Eq.~\eqref{eq:observ}
    \STATE Clip $\bm{\rho}_{k}$ to $[0,\hat \pi_{0}]$
    %\STATE Denote $\bm{\hat\Pi}_{k-1}$ the sigma-points of $\bm{\hat\pi}_{k-1}$
    %\STATE Clip $\bm{\hat\Pi}_{k}$ to $[0, \hat \pi_{0}]$
    %\STATE Transform $\bm{\hat\Pi}$ through the state-transition function to get $\bm{\hat\Pi}_{k+1}^{-}$
    %\STATE Clip $\bm{\hat\Pi}_{k+1}^{-}$ to $[0, \hat \pi_{0}]$
    \STATE Compute $\bm{\hat\pi}_{k+1}$ using UKF with $\bm{\rho}_{k}$ and $\bm{\hat\pi}_{k}$
    \STATE	 $k \gets k+1$
\ENDWHILE
\end{algorithmic}
\end{algorithm}

\subsubsection{Recursive Bayesian Estimation}
Given our linear model Eq.~\eqref{eq:trans_fn}, we wish to compute an optimal estimate of $\bm\pi$ by the conditional expectation,
\begin{equation}
  \label{eq:cond_mean}
\bm{\hat\pi}_{k} = \mathbb{E}[\bm{\pi}_{k} | \bm{\rho}_{0:k}],
\end{equation}
\noindent
where $\bm{\hat\pi}_k$ is the optimal estimate of $\bm{\pi}_{k}$ given the sequence of observations $\bm{\rho}_{0}$ to $\bm{\rho}_{k}$, which we denote $\bm{\rho}_{0:k}$. Note that, Eq.~\eqref{eq:cond_mean} requires one to compute the posterior probability density function (PDF) $p(\bm{\pi}_{k}|\bm{\rho}_{0:k})$, 
\begin{equation}
  \label{eq:aposteriori}
  p(\bm{\pi}_{k}|\bm{\rho}_{0:k}) = \frac{p(\bm{\pi}_{k}|\bm{\rho}_{0:k-1})\cdot p(\bm{\rho}_{k}|\bm{\pi}_{k})}{p(\bm{\rho}_{k}|\bm{\rho}_{0:k-1})},
\end{equation}
\noindent where,
\begin{equation}
  \label{eq:prior}
  p(\bm{\pi}_{k}|\bm{\rho}_{0:k-1}) = \int p(\bm{\pi}_{k}|\bm{\pi}_{k-1}) \cdot p(\bm{\pi}_{k-1}|\bm{\rho}_{0:k-1}) d\bm{\pi}_{k-1},
\end{equation}
\noindent
is a recursive expression of the state at time $k$ as a function of time $k-1$ and the most recent observations.
%The bottom term of Eq. \ref{eq:aposteriori} is a normalization factor that writes:

%\begin{equation}
%  \label{eq:norm_cst}
%  p(\bm{\rho}_{k}|\bm{\rho}_{0:k-1}) = \int p(\bm{\pi}_{k}|\bm{\rho}_{0:k-1}) \cdot p(\bm{\rho}_{k}|\bm{\pi}_{k}) d\bm{\tilde\pi}_{k}
%\end{equation}

%Our state-space model provides expressions of the state-transition probability $p(\bm{\tilde\pi}_{k}|\bm{\tilde\pi}_{k-1})$ via Eq. \ref{eq:trans_fn}, and likelihood $p(\bm{\rho}_{k}|\bm{\tilde\pi}_{k})$ via Eq. \ref{eq:proc_fn}.

Typically, one distinguishes two phases during recursive Bayesian filtering: prediction and correction phase. In the prediction phase, we compute the a-priori state density (Eq. \ref{eq:prior}) using the transition function (Eq. \ref{eq:trans_fn}).
In the correction phase, a new observation vector is available to compute the likelihood $p(\bm{\rho}_{k}|\bm{\tilde\pi}_{k})$ and normalization constant, whereby allowing the a-posteriori state estimate to be computed using Eq.~\eqref{eq:aposteriori}.

Modeling the states as a multi-variate gaussian random variable with additive gaussian noise greatly simplifies this computation, especially when assuming that the transition and observation models are linear. In particular, the latter assumptions typically allow for Kalman Filter type solutions~\citep{kalman1960}. However, the present scenario imposes an inequality constraint on the states so as to make them interpretable as probabilities, a requirement that standard Kalman Filters does not allow for. Instead,~\cite{gupta07} suggests applying an intermediate step in which the a-priori state estimates are projected on the constraint surface. This approach, despite being effective, requires the solving a quadratic program at each iteration.
\begin{figure*}[t]
\caption{Visualization of our SSnnPU method on an example MRI volume. (Top row): Single slice from volume with groundtruth segmentation highlighted (red) and a point-wise annotation (green), output prediction at different epochs. (Bottom row): Class priors at corresponding epochs over 75 slices of the volume. The different curves correspond to: the true priors (blue), observations $\rho_{k}$ (orange), proportion of pseudo-positives (green), and current estimated priors $\hat\pi_{k}$ (red). The stopping condition is triggered at epoch 71. {Figure best seen in color.}}
\centering
    \includegraphics[width=.9\textwidth]{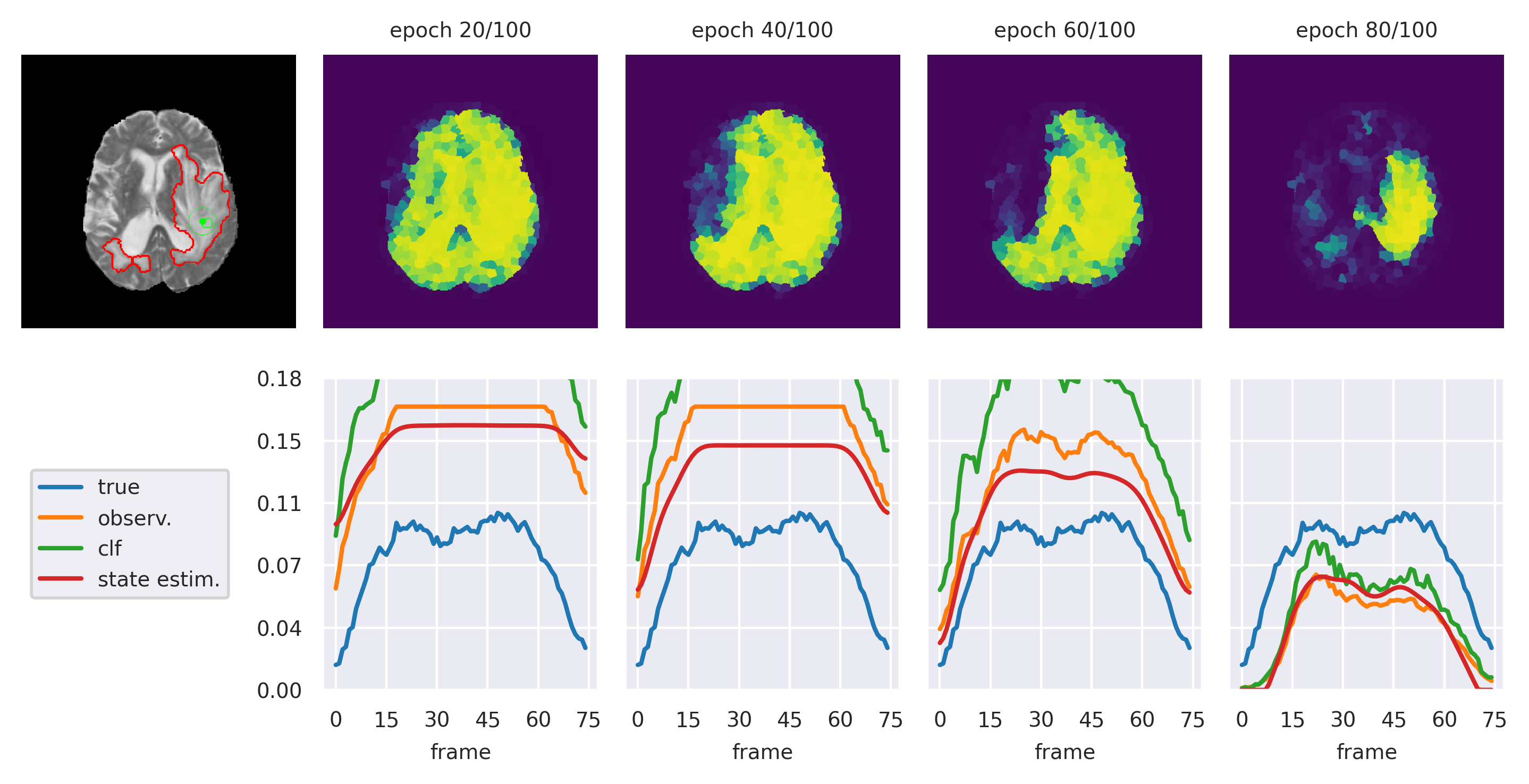}
\label{fig:prevs_conv}
\end{figure*}

Instead, we use the simpler approach of~\cite{kandepu08}, which relies on the Unscented Kalman Filter (UKF) approach~\citep{wan00}. In contrast with standard Kalman Filters, which propagate the means and covariances of states through the (linear) system, UKF samples a set of carefully chosen points from the state distribution, called {\it sigma-points}, that allow to accurately approximate the true statistics of the posterior. Our inequality constraints are then directly applied to the sigma-points.
% TODO: introduce capital pi

As illustrated in Fig.~\ref{fig:prevs_conv}, our self-supervised approach iteratively decreases its estimates of $\pi^i$ (red line) as a function of observations $\rho_{k}^{i}$ (orange line). Given the defined control inputs $u_k$ which induce a downward acceleration, $\pi^i$ are reduced further than they should be (see epoch 80 in Fig.~\ref{fig:prevs_conv}). For this reason, we introduce a strategy to halt this computation at an appropriate moment in the next section.

\subsubsection{Stopping condition}
We introduce two stopping conditions that we use simultaneously to ensure that our methods stops at an appropriate value for $\bm\hat\pi$.

Specifically, first we denote $\bm{\tilde{\mathcal{X}}}_{n}=\{x \in \bm{\mathcal{X}} | f_{\theta}(x) < 0.5\}$, and $\bm{\tilde{\mathcal{X}}}_{p}=\{x \in \bm{\mathcal{X}} | f_{\theta}(x) \geq 0.5\}$ as the set of ``pseudo-negative'' and ``pseudo-positive'' samples, respectively.
As a first criteria, we use the variance of the predictions of  $\bm{\tilde{\mathcal{X}}}_{n}$, written $Var[f_{\theta}(\bm{\mathcal{\tilde X}}_{n})]$, which measures the confidence of our model on the negative samples (\ie background regions of the image).
Second, we also impose that our predictions are such that the frequency of positives is below our upper-bound on all frames. By combining these, our criteria (1) imposes a maximum on the frequencies of pseudo-positives, (\ie $\frac{|\bm{\tilde \mathcal{X}}_{p}^{i}|}{N_{i}} < \hat \pi_{0} \quad \forall i$) and (2) imposes a maximum variance level to pseudo-negatives (\ie $Var[f_{\theta}(\bm{\mathcal{\tilde X}}_{n})] < \tau$).

In practice, we want the above conditions to be verified for several iterations so as to guarantee stability. We therefore impose that both conditions are satisfied for $T_{s}$ iterations. In Fig.~\ref{fig:prevs_conv}, we illustrate the behaviour of our stopping conditions by showing the predicted probabilities and their corresponding class-priors on a given sequence.

%%% Local Variables:
%%% mode: latex
%%% TeX-master: "main"
%%% End:

%% file: method_tracking.tex
\subsection{Spatio-Temporal Regularization}
\label{sec:tracking}

While our \SSnnPU~method leverages all images and point-wise annotations to train and segment the data volume in question, the output of our method does not explicitly leverage the spatio-temporal relations within the data cube. That is, every sample is treated and predicted independently, and only implicitly related through $f_\theta$. In order to coherently regularize over the different frames and locations, we use an existing graph based framework as a post-processing step.

To this end, we make use of the multi-object tracking framework (\KSPTrack) introduced in~\cite{lejeune18} to refine the output of the \SSnnPU~method. In short, \KSPTrack~represents the data volume with superpixels and builds a network graph over these to optimize a set of spatio-temporal paths that jointly correspond to the object segmentation throughout the data volume. This is solved by casting the problem as network-flow optimization, whereby costs are assigned to input/output nodes, visiting and transition edges within and across frames, and where 2D annotations are used to define source nodes that allow to push flow within the network.

In practice, we use the same orginal \KSPTrack~setup as in~\cite{lejeune18} with the exception of using the output of \SSnnPU, $f_{\theta}(x_i)$ to compute the cost of selecting superpixel $x_i$ as part of the object by,
\begin{equation}
  \label{eq:cost_fg}
  C_{fg}(i) = -\log \frac{f_\theta(x_i)}{1-f_\theta(x_i)},
\end{equation}
\noindent
where $C_{fg}(i)$ is the cost of including superpixed $x_i$ as part of the object. By construction, this relation therefore imposes a negative cost when $f_{\theta}(x) > 0.5$, and a non-negative cost otherwise. 

The final output of the \KSPTrack~method yields a binary image for each of the frames in the data volume. For the remainder of this paper, we will refer to the combined use of \SSnnPU~and \KSPTrack~as \SSnnPUKSP.

%%% Local Variables:
%%% mode: latex
%%% TeX-master: "main"
%%% End:

%% file: method_train_details.tex
\subsection{Training details, hyper-parameters, and implementation}
\label{sec:implementation}
We now specify technical details of our implementation and training procedure. \SSnnPU~is implemented in Python using the PyTorch library\footnote{\red{\url{https://github.com/lejeunel/ssnnpu_ksptrack}}}, while we use a publicly available \red{C++} implementation of \red{the K-shortest paths algorithm for the spatio-temporal regularization step}\footnote{\red{\url{https://github.com/lejeunel/pyksp}}}.

\subsubsection{SSnnPU}
$f_\theta$ is implemented as a Convolutional Neural Network based on the U-Net architecture proposed in~\cite{ibtehaz20} for all experiments. It uses ``Inception-like'' blocks in place of simple convolutional layers to improve robustness to scale variations. Skip connections are replaced by a serie of $3\times3$ convolutional layers with residual connections. Batch normalization layers are added after each convolutional layer.

To train \SSnnPU, we proceed with a three phase process:
\begin{enumerate}
  \item To increase the robustness of early observations, we train $f$ for 50 epochs with Alg.~\ref{alg:sgdnnpu} and a learning rate set to $10^{-4}$. With the last layer of our decoder being a sigmoid function, we set the bias of the preceding convolutional layer to $-\log{\frac{1-\pi_{init}}{\pi_{init}}}$, with $\pi_{init}=0.01$, as advised in~\cite{lin17}. All others parameters are initialized using He's method~\cite{he15init}.
    \item We then optimize the model and class-prior estimates for a maximum $100$ epochs as described in Alg.~ \ref{alg:prior_estim} with a learning rate set to $10^{-5}$.
  \item We then train using frame-wise priors given by the previous phase for an additional $100$ epochs with a learning rate of $10^{-5}$.
\end{enumerate}
We use the Adam optimizer with weight decay $0.01$ for all training phases. Data augmentation is performed using a random combination of additive gaussian noise, bilateral blur and gamma contrasting. 

\subsubsection{Recursive Bayesian Estimation}
For the process, transition, and initial covariance matrices, we use diagonal matrices 
$Q=\sigma_{Q}\mathbb{I}$, $R=\sigma_{R}\mathbb{I}$, and $S=\sigma_{S}\mathbb{I}$, where $\mathbb{I}$ is the identity matrix.
As the observations $\rho_{k}^{i}$ are often very noisy, we set $\gamma=2$ and the observation variance much larger than the process variance $\sigma_{Q}=10$, $\sigma_{R}=0.05$ and $\sigma_{S}=0.03$.
The parameters of the control input are set proportionally to $\hat \pi_{0}$ with $u_{0}=0.02 \hat \pi_{0}$, and $u_{T}=0.4 \hat \pi_{0}$. The window length of the frame-wise smoothing filter is set proportionally to the number of frames: $L=0.05N$. The time-period of our stopping condition is set to $T_{s}=10$ and the threshold on the variance is $\tau=0.007$.

\subsubsection{KSPTrack parameters}
\label{sec:org4560526}
All sequences are pre-segmented into $\sim 1200$ superpixels and the output of $f$ is averaged over all pixels in a superpixel. Each point-wise annotation defines a circle of radius \(R=0.05 \cdot \max\{W,H\}\) centered on the 2D location, where \(W\) and \(H\) are the width and height of frames, respectively. The input cost at given superpixel is set to $0$ when its centroid is contained within that circle, and $\infty$ otherwise. The transitions costs are set to $0$ when superpixels overlap and $\infty$ otherwise.
In order to reduce the number of edges and alleviate the computational requirements, we also prune {visiting} edges when their corresponding object probability falls below $0.4$. We perform a single round of \KSPTrack~as augmenting the set of positives and re-training the object model after each round (as in~\cite{lejeune18}) did not prove beneficial.

%%% Local Variables:
%%% mode: latex
%%% TeX-master: "main"
%%% End:

%% file: exp.tex
\section{Experiments}
\label{sec:exps}

In the following section, we outline the experiments performed to characterize the behavior of our method. First, we compare our method with existing baselines for segmentation purposes. We then perform an ablation study to demonstrate which aspect of our method provides what quantitative benefits, as well as the impact of the class-prior upper-bound initialization. Last, we show how our stopping condition performs in establishing useful class priors.

\subsection{Datasets}
\label{sec:datasets}
To validate our method, we evaluate it on the publicly available dataset used in~\cite{lejeune18}\footnote{\red{Datasets, manual ground truth annotations, and point-wise annotations used in this paper are available at \url{https://doi.org/10.5281/zenodo.5007788}}}.
It consists of a variety of video and volumes of different modalities with 2D annotation points for different objects of interest, as well as the associated groundtruth segmentations. Specifically, it includes four different subsets of data:
\begin{itemize}
\item \textbf{Brain:} Four 3D T2-weighted MRI scans chosen at random from the publicly available BRATS challenge dataset~\citep{menze15}, where tumors are the object of interest.
\item \textbf{Tweezer:} Four sequences from the training set of the publicly dataset MICCAI Endoscopic Vision challenge: Robotic Instruments segmentation. The surgical instrument is the object to segment in these sequences.
\item \textbf{Slitlamp:} Four slit-lamp video recordings of human retinas, where the optic disk is to be segmented.
\item \textbf{Cochlea:} Four volumes of 3D CT scans of the inner ear, where the cochlea must be annotated. This object is the most challenging object to segment due to its challenging geometry (\ie non-concave shape).
\end{itemize}

\subsection{Baselines and experimental setup}
Using the datasets mentioned above, we evaluate the proposed methods (\SSnnPU~and \SSnnPUKSP) quantitatively and qualitatively.
Additionally, we compare these to existing baseline methods that perform the same tasks. These include:
\begin{itemize}
\item \KSPTrack: Multi-object tracking method described in \cite{lejeune18}. As in the original work, the object model consists of a decision tree bagger adapted to the PU setting, while features are taken from a CNN configured as an autoencoder.
\item \textbf{EEL}: An expected exponential loss within a boosting framework for robust classication in a PU learning setting~\citep{lejeune17}.
\item \textbf{Gaze2Segment}: A learned saliency-map based detection regularized with graphcut~\citep{khosravan16}.
\item \textbf{DL-prior}: Point location supervision is used to train a CNN with strong object priors~\citep{bearman16}.
  \item \red{\textbf{AlterEstPU}: An alternating training and class-prior estimation method inspired by the expectation-maximization algorithm that leverages the non-negative Positive/Unlabeled risk estimator of~\cite{kiryo17} as described in~\cite{kato18}. This uses the same model, training scheme and parameters as~\SSnnPU~(Sec.~\ref{sec:implementation}), with the exception that the second phase is replaced by the class-prior update scheme of~\cite{kato18}. Note that in contrast to~\SSnnPU, a single value for the class-prior is estimated. We tune the parameters of the update scheme for best performance on the tested sequences.}
\end{itemize}

Methods \textbf{ELL} and \textbf{KSPTrack} have been specifically designed for the proposed evaluation datasets (sec. \ref{sec:datasets}), and their parameters have been optimized for best performance. Methods \textbf{Gaze2Segment} and \textbf{DL-prior} have been implemented and optimized to give their best performances.

For our method, we set the initial class prior upper-bound, $\pi_{max}$, by computing the frequencies of positives $\pi^{i}$ for all frames from the groundtruth and set $\hat \pi_{0}=1.4 \cdot \max_{i} \{ \pi^{i} \}$. While this may appear to be using the groundtruth to set the parameters of our method, it allows us to calibrate $\pi$ for comparison reasons. Indeed, a 40\% factor over the frame that has the largest object surface object can represent the entire image depending on the dataset. To show that our method is not inherently sensitive to this value, we perform an analysis of sensitivity in Sec.~\ref{sec:ablation}.

Following a transductive learning setup, the input of each of evaluated methods is a single data volume and their associated 2D point annotations, and yields a complete segmentation for the inputted data volume.

\input{tables/results.tex}
\begin{figure*}[h]
\caption{Qualitative results for each type. From left to right: Original image with groundtruth highlighted in red and 2D location in green, output of foreground prediction model (SSnnPU), SSnnPU combined with our spatio-temporal regularization scheme (SSnnPU+KSPTrack), best baseline (KSPTrack), second best baseline (AlterEstPU), and third best baseline (EEL).}
\centering
    \includegraphics[width=.95\textwidth]{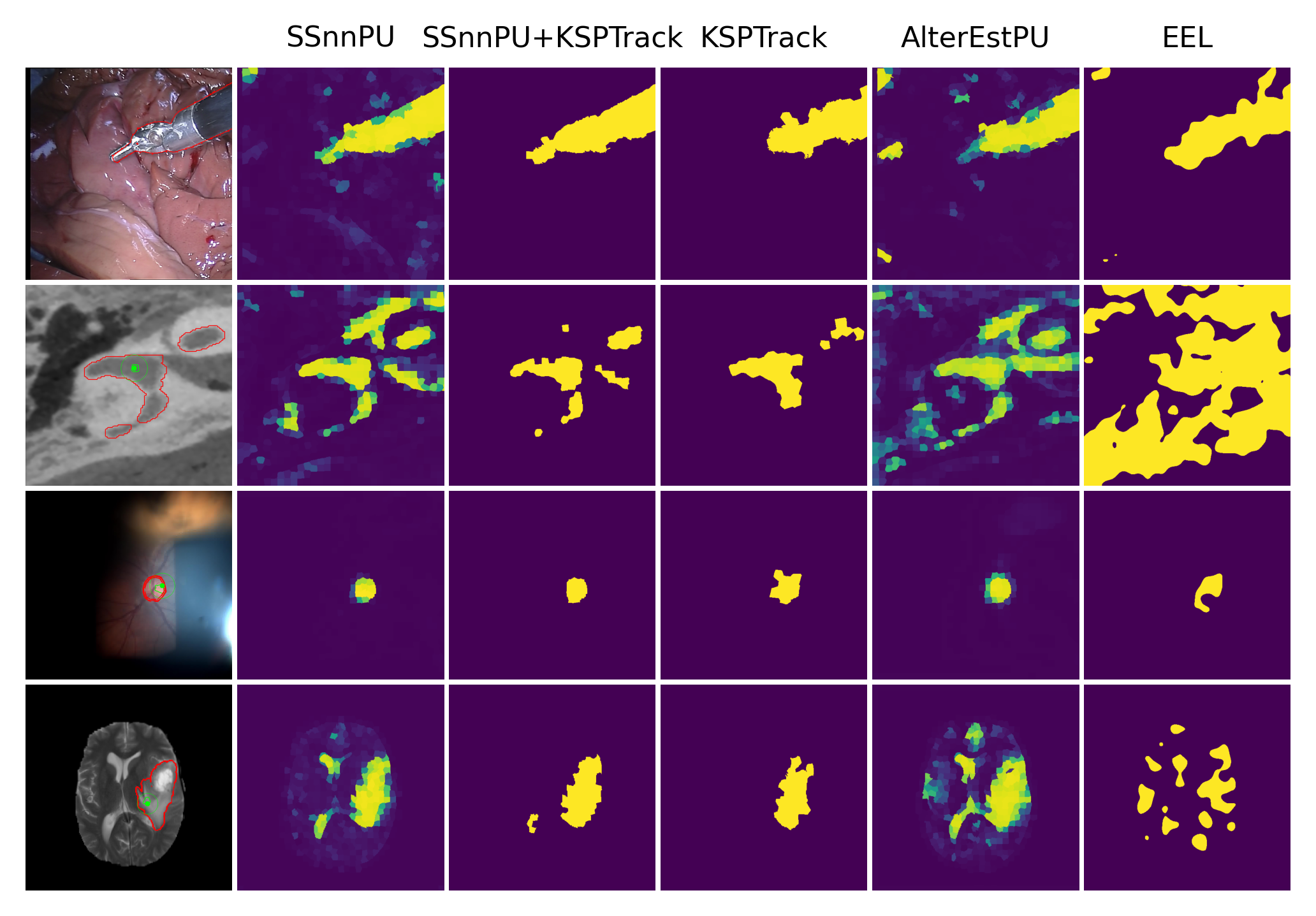}
\label{fig:prevs}
\end{figure*}

\begin{figure*}[t]
\caption{Visualization of stopping conditions for \SSnnPU~method. For each tested image type, we show: (top) Mean Absolute Error (MAE) between the estimated and the true prior, (bottom) Variance of the pseudo-negatives. We also show here threshold (in dashed-red), and the minimum number of epochs (dashed-green). Both the optimal (circle) and the stopping conditions-based class-priors (cross) are shown on each of the sequences of each type (one line per sequence). }
\centering
\begin{tabular}{@{}cc@{}}
    \includegraphics[width=.5\textwidth]{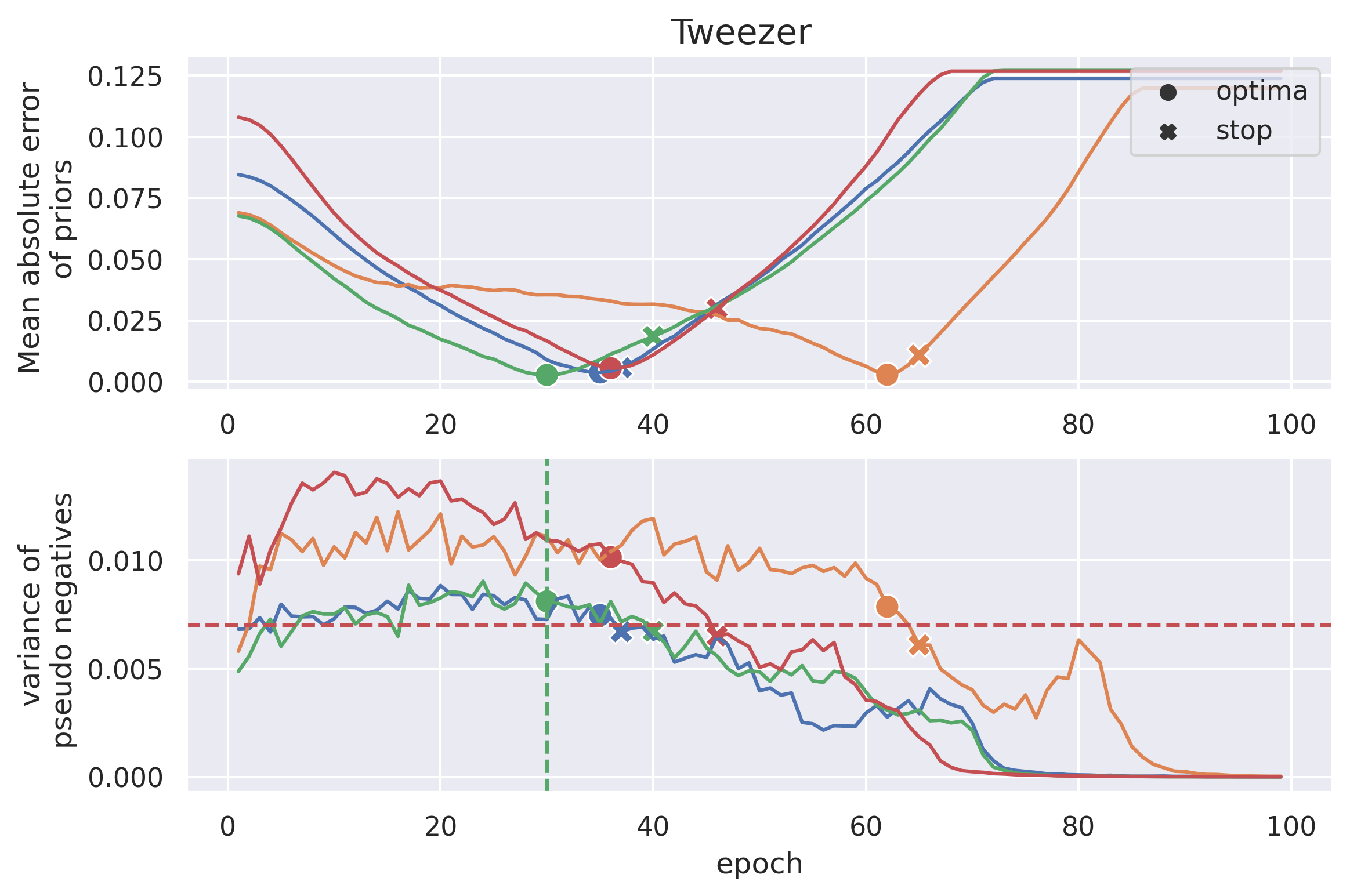} &
    \includegraphics[width=.5\textwidth]{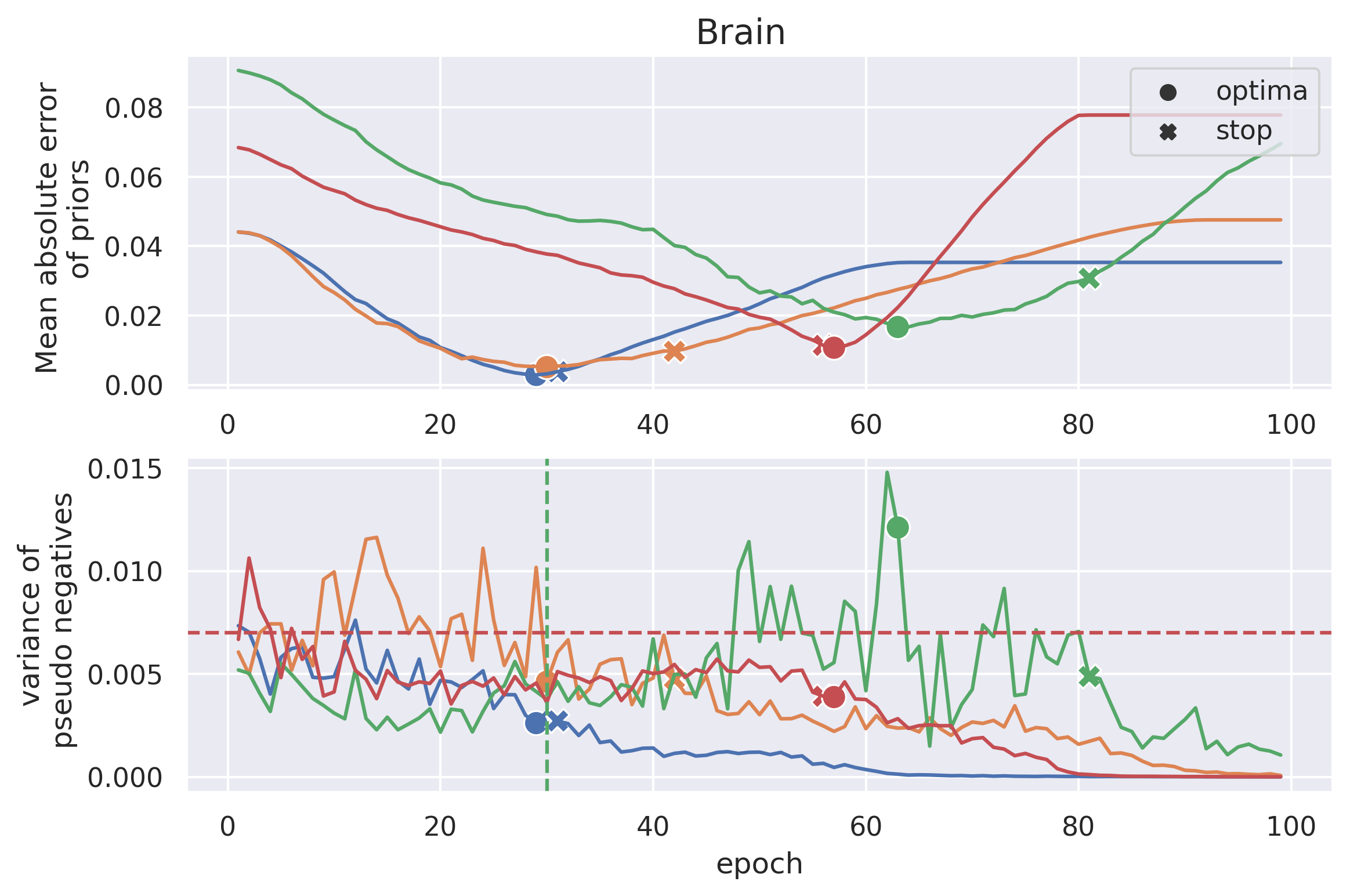} \\
    \includegraphics[width=.5\textwidth]{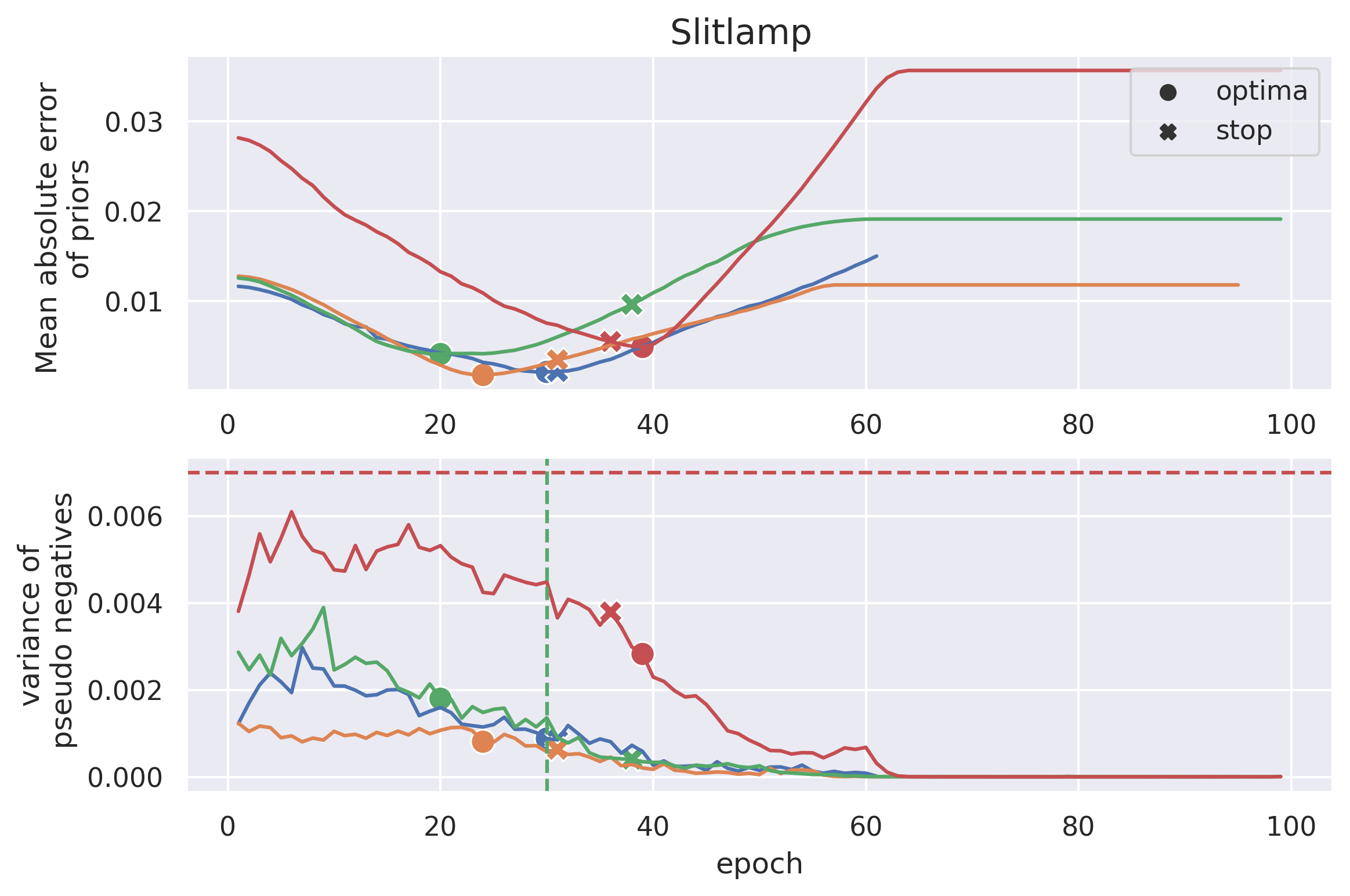} &
    \includegraphics[width=.5\textwidth]{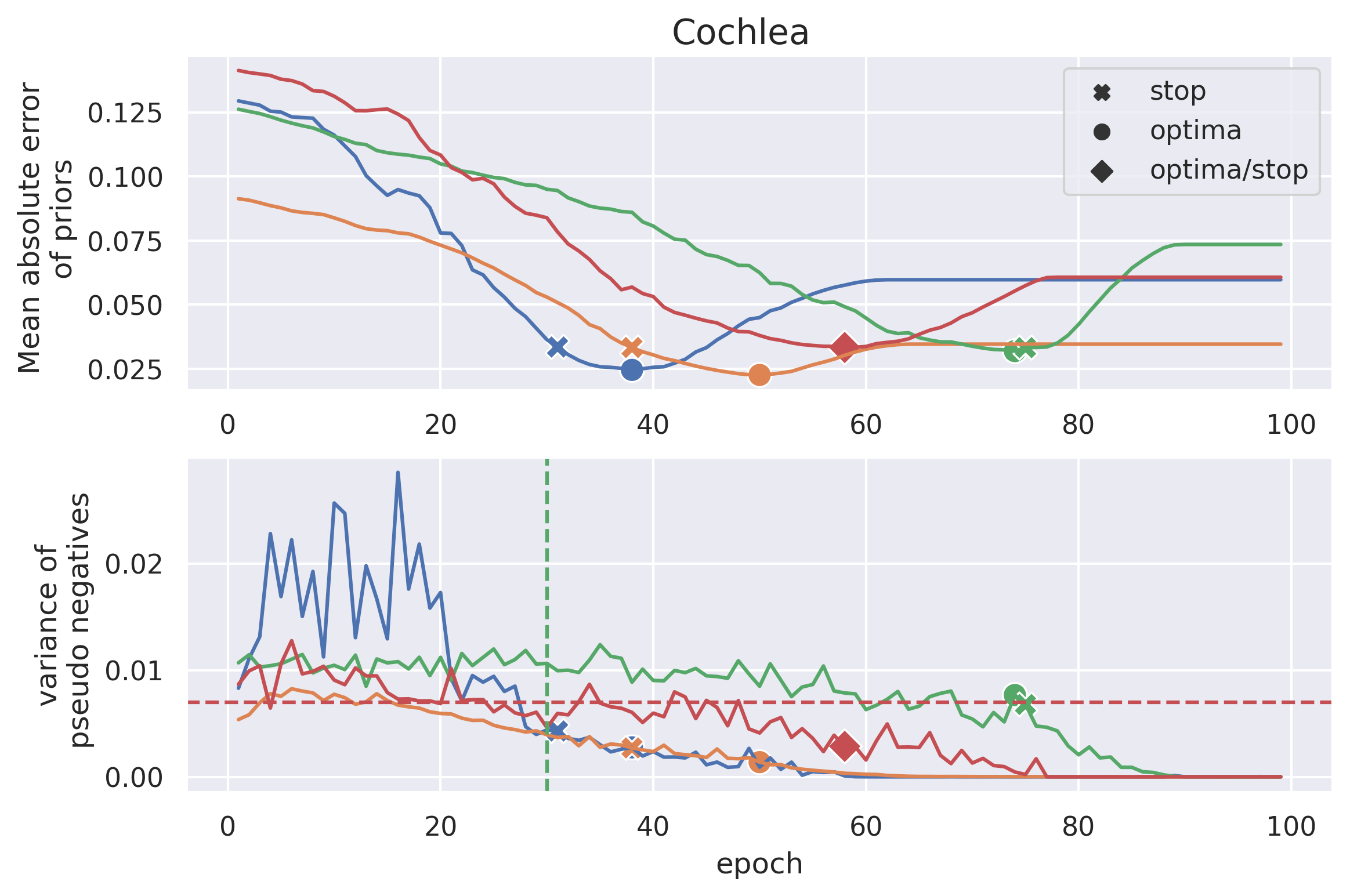} \\
\end{tabular}
\label{fig:converge}
\end{figure*}

\subsection{Segmentation performance}
\label{sec:exp_performance}

In Table.~\ref{tab:results}, we show the F1 scores of each method, averaged over each sequence for each data type. Given that some methods require superpixels, we also show the maximum performance a segmentation method would have if every single superpixel was correctly labeled. In practice we denote positive superpixels as those with more than half of the pixels are in the groundtruth. We then compare the latter segmentation with the pixel-wise manual ground truth annotation and denote the result \MaxSP.

From these results, we note that both \SSnnPU~and \SSnnPUKSP~ perform well on average. Most notably, \SSnnPUKSP~outperformes all other approaches including \SSnnPU. This is coherent as the spatio-temporal regularization provides an efficient method to remove false positives generated by \SSnnPU. On the Tweezer sequences, the gain in performances are striking, with an increase in 14\% on average over the previous state-of-the-art, and closely approaches a perfect labeling according to \MaxSP.

When comparing \SSnnPU~and \SSnnPUKSP, we note that the spatio-temporal regularization provided by the KSP framework is particularly effective in the Cochlea cases (\ie plus 18\%). This is coherent as the geometry of the cochlea is largely made of a rings and where visual appearance plays a somewhat lesser role in identifying the complete structure. This latter point also explains the fairly poor performance of \SSnnPU~but much improved one by \SSnnPUKSP~when compared to \KSPTrack.

\red{Last, comparing \SSnnPU~and \AlterEstPU, we observe important limitations of the latter method, which, aside from the fact that it only estimates a single value for the class-prior over the whole sequence, tends to be very sensitive to noise and often fails to converge, thereby showing inferior performance and higher variance.}

In Fig.~\ref{fig:prevs}, we show qualitative results of different methods and provide complete video results of \SSnnPUKSP~ as supplementary material.

\subsection{Ablation study}
\label{sec:ablation}
To provide a better understanding as to what aspects of our method provide improvements, we perform an ablation study. Specifically, we evaluate the following variants of our methods:
\begin{itemize}
\item \SSnnPU: Non-negative positive-unlabeled with self-supervised class-prior estimation, as in Sec.~\ref{sec:nnpu} and~\ref{sec:pi_estim}.
\item \SSnnPUKSP: Combines the \SSnnPU~method and the \KSPTrack~methods described in Sec.~\ref{sec:tracking}.
\item \SSnnPUTrue: Same as \SSnnPUKSP, except that the foreground model is directly trained using the true class-priors (\ie taken from the groundtruth) following Alg.~\ref{alg:sgdnnpu}.
\item \SSnnPUConst: Same as \SSnnPUTrue, except that we use a sequence-wise constant class prior given by the mean groundtruth prior over all frames (as in~\cite{kiryo17}).
\end{itemize}
For both \SSnnPUTrue~and \SSnnPUConst, we train the models $f$ for $150$ epochs. The learning rate in both cases are set to $10^{-4}$ and reduced to $10^{-5}$ after $50$ epochs.

\input{tables/table_eta.tex}

In addition, to assess how \SSnnPUKSP~performs as a function of the selected $\pi_{max}$, we evaluate the method using $\hat \pi_{0}=\eta \max_{i}\pi^{i}$, with $\eta \in \{1.2, 1.4, 1.6, 1.8\}$. Similarly, to assess the relevance of the self-supervised estimation of priors, we perform segmentation using method \SSnnPUConst~with $\eta \in \{0.8, 1.0, 1.2, 1.4\}$.

We report F1, precision and recall scores for each aforementioned method in Table.~\ref{tab:results_eta}. First, we observe that our self-supervised estimation is relatively robust to variations in $\hat \pi_{0}$.
In particular, Tweezer shows variations in F1 scores of at most $1\%$ while $\eta$ ranges between $1.2$ and $1.6$. Similarly, the performances fluctuate only marginally for Cochlea, Slitlamp and Brain sequences with at most 5\% changes. These fluctuations still lead to improved performances over state-of-the-art methods.

In contrast, sensitivity to initial class priors is much stronger for \SSnnPUConst, which yields variations between 1\% and 15\% depending on the sequence type. The relevance of the frame-wise prior estimation can also be assessed by comparing, for each sequence type, the maximum F1 scores reached by \SSnnPUKSP~and \SSnnPUConst~for all tested values of $\eta$. \SSnnPUKSP~brings an improvement over \SSnnPUConst~of $2\%, 1\%, 6\%$ and $0\%$ for the Tweezer, Brain, Cochlea and Slitlamp sequence, respectively.

Last, comparing \SSnnPUKSP~to \SSnnPUTrue, we note that the proposed self-supervised prior estimation framework brings comparable performances on all types. That is the \SSnnPU~ component of our methods appears to provide class priors on par with the true class-priors. Surprisingly, in the Tweezer case, some values of $\eta$ yield even better performances than if the true class prior values were to be used.

\subsection{Analysis of Convergence}
\label{sec:convergence}

Last, we analyze the behavior of our proposed stopping conditions in Alg.~\ref{alg:prior_estim}. In particular, Fig.~\ref{fig:converge} illustrates the convergence of the class prior estimation for each type of sequence. In these cases, all sequences are trained using $\hat\pi_{0}= 1.4 \cdot \max_i \{\pi^i\}$.

As we aim to estimate the true class-prior, we plot the Mean Absolute Error between the estimated priors and the true priors (top panels).
The proposed stopping condition, which leverages the variance of pseudo-negatives, is plotted in the bottom panel. We observe that the proposed stopping condition triggers reasonably close to the optima in most cases. In cases where the stopping condition is triggered earlier or later, we note that the difference in mean absolute error is fairly small, whereby implying that the impact in the early/late trigger is rather small.

%%% Local Variables:
%%% mode: latex
%%% TeX-master: "main"
%%% End:

%% file: tables/results.tex
\begin{table*}[t]
\centering
\caption{
    Quantitative results on all datasets. We report the F1 scores and standard deviations.
    In column ``All'', we show the average F1 score on all sequences.
    In column $\Delta$, we show the absolute difference with respect to the maximum achievable score given the superpixel over-segmentation (Max. SP).
    }
\label{tab:results}
\begin{tabular}{llp{1.8cm}p{1.8cm}p{1.8cm}p{1.8cm}p{1.8cm}}
\toprule
Types &              Tweezer &              Cochlea &             Slitlamp &                Brain &          All &      $\Delta$ \\
Methods         &                      &                      &                      &                      &              &               \\
\midrule
Max. SP         &       $0.92\pm 0.02$ &       $0.92\pm 0.01$ &       $0.92\pm 0.02$ &       $0.95\pm 0.01$ &       $0.93$ &             - \\
\hline

SSnnPU+KSPTrack &  $\bm{0.91}\pm 0.03$ &  $\bm{0.75}\pm 0.05$ &  $\bm{0.84}\pm 0.05$ &  $\bm{0.80}\pm 0.09$ &  $\bm{0.82}$ &  $\bm{-0.10}$ \\
SSnnPU          &       $0.87\pm 0.02$ &       $0.53\pm 0.10$ &       $0.78\pm 0.10$ &       $0.75\pm 0.13$ &       $0.73$ &       $-0.19$ \\
\hdashline
KSPTrack        &       $0.77\pm 0.08$ &       $0.66\pm 0.02$ &       $0.77\pm 0.08$ &       $0.74\pm 0.08$ &       $0.74$ &       $-0.19$ \\
AlterEstPU        &       $0.74\pm 0.12$ &       $0.39\pm 0.11$ &       $0.65\pm 0.11$ &       $0.53\pm 0.24$ &       $0.58$ &       $-0.35$ \\
EEL             &       $0.60\pm 0.16$ &       $0.12\pm 0.05$ &       $0.59\pm 0.08$ &       $0.52\pm 0.14$ &       $0.46$ &       $-0.47$ \\
Gaze2Segment    &       $0.18\pm 0.00$ &       $0.07\pm 0.02$ &       $0.02\pm 0.00$ &       $0.07\pm 0.02$ &       $0.08$ &       $-0.84$ \\
DL-prior        &       $0.72\pm 0.06$ &       $0.30\pm 0.04$ &       $0.51\pm 0.11$ &       $0.56\pm 0.08$ &       $0.52$ &       $-0.40$ \\
\bottomrule
\end{tabular}
\end{table*}

%% file: tables/table_eta.tex
\begin{table*}
\centering
\caption{
    Quantitative results on all datasets for different prior levels. We report the F1 score, precision (PR), recall(RC) and standard deviations.
    }
\label{tab:results_eta}
\begin{tabular}{llp{1.8cm}p{1.8cm}p{1.8cm}p{1.8cm}}
\toprule
      &                   &   &              F1 &              PR &              RC \\
Types & Methods & $\eta$ &                 &                 &                 \\
\midrule
\multirow{9}{*}{Tweezer} & \multirow{4}{*}{nnPUconst+KSPTrack} & 0.8 &  $0.89\pm 0.03$ &  $0.91\pm 0.05$ &  $0.88\pm 0.03$ \\
      &                   & 1.0 &  $0.90\pm 0.04$ &  $0.86\pm 0.06$ &  $0.93\pm 0.02$ \\
      &                   & 1.2 &  $0.90\pm 0.03$ &  $0.84\pm 0.05$ &  $0.96\pm 0.01$ \\
      &                   & 1.4 &  $0.89\pm 0.03$ &  $0.83\pm 0.05$ &  $0.97\pm 0.00$ \\
\cline{2-6}
      & \multirow{4}{*}{SSnnPU+KSPTrack} & 1.2 &  $0.92\pm 0.01$ &  $0.93\pm 0.02$ &  $0.92\pm 0.03$ \\
      &                   & 1.4 &  $0.91\pm 0.03$ &  $0.91\pm 0.03$ &  $0.91\pm 0.05$ \\
      &                   & 1.6 &  $0.91\pm 0.03$ &  $0.88\pm 0.03$ &  $0.94\pm 0.03$ \\
      &                   & 1.8 &  $0.90\pm 0.04$ &  $0.88\pm 0.04$ &  $0.92\pm 0.04$ \\
\cline{2-6}
      & nnPUtrue+KSPTrack & - &  $0.90\pm 0.03$ &  $0.89\pm 0.03$ &  $0.92\pm 0.04$ \\
\cline{1-6}
\multirow{9}{*}{Cochlea} & \multirow{4}{*}{nnPUconst+KSPTrack} & 0.8 &  $0.59\pm 0.15$ &  $0.88\pm 0.09$ &  $0.45\pm 0.15$ \\
      &                   & 1.0 &  $0.67\pm 0.07$ &  $0.87\pm 0.14$ &  $0.55\pm 0.06$ \\
      &                   & 1.2 &  $0.71\pm 0.10$ &  $0.91\pm 0.10$ &  $0.59\pm 0.12$ \\
      &                   & 1.4 &  $0.68\pm 0.10$ &  $0.77\pm 0.18$ &  $0.62\pm 0.11$ \\
\cline{2-6}
      & \multirow{4}{*}{SSnnPU+KSPTrack} & 1.2 &  $0.73\pm 0.08$ &  $0.85\pm 0.16$ &  $0.65\pm 0.03$ \\
      &                   & 1.4 &  $0.75\pm 0.05$ &  $0.88\pm 0.08$ &  $0.66\pm 0.04$ \\
      &                   & 1.6 &  $0.72\pm 0.07$ &  $0.80\pm 0.20$ &  $0.68\pm 0.07$ \\
      &                   & 1.8 &  $0.64\pm 0.14$ &  $0.80\pm 0.32$ &  $0.58\pm 0.07$ \\
\cline{2-6}
      & nnPUtrue+KSPTrack & - &  $0.76\pm 0.05$ &  $0.85\pm 0.08$ &  $0.69\pm 0.03$ \\
\cline{1-6}
\multirow{9}{*}{Slitlamp} & \multirow{4}{*}{nnPUconst+KSPTrack} & 0.8 &  $0.71\pm 0.30$ &  $0.84\pm 0.08$ &  $0.72\pm 0.38$ \\
      &                   & 1.0 &  $0.84\pm 0.03$ &  $0.79\pm 0.05$ &  $0.91\pm 0.04$ \\
      &                   & 1.2 &  $0.78\pm 0.05$ &  $0.66\pm 0.07$ &  $0.95\pm 0.01$ \\
      &                   & 1.4 &  $0.66\pm 0.04$ &  $0.51\pm 0.05$ &  $0.95\pm 0.01$ \\
\cline{2-6}
      & \multirow{4}{*}{SSnnPU+KSPTrack} & 1.2 &  $0.72\pm 0.29$ &  $0.92\pm 0.05$ &  $0.67\pm 0.34$ \\
      &                   & 1.4 &  $0.84\pm 0.05$ &  $0.85\pm 0.08$ &  $0.84\pm 0.12$ \\
      &                   & 1.6 &  $0.80\pm 0.11$ &  $0.89\pm 0.06$ &  $0.75\pm 0.19$ \\
      &                   & 1.8 &  $0.84\pm 0.04$ &  $0.77\pm 0.06$ &  $0.92\pm 0.02$ \\
\cline{2-6}
      & nnPUtrue+KSPTrack & - &  $0.84\pm 0.03$ &  $0.79\pm 0.04$ &  $0.90\pm 0.02$ \\
\cline{1-6}
\multirow{9}{*}{Brain} & \multirow{4}{*}{nnPUconst+KSPTrack} & 0.8 &  $0.79\pm 0.09$ &  $0.82\pm 0.18$ &  $0.78\pm 0.04$ \\
      &                   & 1.0 &  $0.78\pm 0.11$ &  $0.77\pm 0.12$ &  $0.79\pm 0.10$ \\
      &                   & 1.2 &  $0.72\pm 0.09$ &  $0.60\pm 0.10$ &  $0.92\pm 0.06$ \\
      &                   & 1.4 &  $0.73\pm 0.08$ &  $0.60\pm 0.09$ &  $0.93\pm 0.06$ \\
\cline{2-6}
      & \multirow{4}{*}{SSnnPU+KSPTrack} & 1.2 &  $0.79\pm 0.10$ &  $0.82\pm 0.14$ &  $0.76\pm 0.07$ \\
      &                   & 1.4 &  $0.80\pm 0.09$ &  $0.78\pm 0.14$ &  $0.84\pm 0.07$ \\
      &                   & 1.6 &  $0.77\pm 0.10$ &  $0.80\pm 0.12$ &  $0.75\pm 0.12$ \\
      &                   & 1.8 &  $0.76\pm 0.11$ &  $0.80\pm 0.11$ &  $0.73\pm 0.17$ \\
\cline{2-6}
      & nnPUtrue+KSPTrack & - &  $0.80\pm 0.09$ &  $0.73\pm 0.10$ &  $0.89\pm 0.07$ \\
\bottomrule
\end{tabular}
\end{table*}

%% file: conclusion.tex
\section{Conclusions}
\label{sec:conclusion}

The present work contributes to the challenging problem of segmenting medical sequences of various modalities using point-wise annotations and without knowing in advance what is to be segmented. As such, it has important implications in the ability to quickly produce groundtruth annotations or learn from a single example.

By formulating our problem as a positive/unlabeled prediction task, we demonstrated the relevance of the non-negative unbiased risk estimator as a loss function of a Deep Convolutional Neural Network. Our novel contribution of a self-supervised framework based on recursive bayesian filtering, to estimate the class priors, a hyper-parameter that plays an important role in the segmentation accuracy, was demonstrated to bring important performance gains over state-of-the-art methods, particularly when also used in combination with a spatio-temporal regularlization scheme. From the annotator's point of view, the burden is marginally increased in what the method requires, asides from 2D locations and an upper-bound on the class-prior. While our approach does not perform flawlessly in challenging cases, we show that the performance is stable and resilient to miss-specified class prior upper-bounds. Last, we show that our stopping conditions are adept at yielding favorable estimates as well.

As future works, we aim at investigating the addition of negative samples into our foreground model, by leveraging the positive, unlabeled, and biased negative risk estimator of~\cite{hsieh19}. Similarly, given that our regularization is outside of the learning framework, we wish to also explore how this could be integrated to increase performance. \red{Last, there are interesting potentials in using recent graph-based PU learning approaches~\cite{akujuobi2020} to see how additional data can be sequentially added and annotated. In this way, trained models per sequence could be aggregated as more data is available. }

%%% Local Variables:
%%% mode: latex
%%% TeX-master: "main"
%%% End: